\documentclass[twoside,twocolumn,9pt]{article}

\usepackage{extsizes}
\usepackage[super,sort&compress,comma]{natbib} 
\usepackage[version=3]{mhchem}
\usepackage[left=1.5cm, right=1.5cm, top=1.785cm, bottom=2.0cm]{geometry}
\usepackage{balance}
\usepackage{mathptmx}
\usepackage{sectsty}
\usepackage{graphicx} 
\usepackage{lastpage}
\usepackage[format=plain,justification=justified,singlelinecheck=false,font={stretch=1.125,small,sf},labelfont=bf,labelsep=space]{caption}
\usepackage{float}
\usepackage{fancyhdr}
\usepackage{fnpos}
\usepackage{fixltx2e}
\usepackage[english]{babel}
\addto{\captionsenglish}{%
  
}
\usepackage{multirow}
\usepackage{array}
\usepackage{droidsans}
\usepackage{charter}
\usepackage[T1]{fontenc}
\usepackage[usenames,dvipsnames]{xcolor}
\usepackage{setspace}
\usepackage[compact]{titlesec}
\usepackage{hyperref}
\usepackage{algorithm}
\usepackage{algpseudocode}

\usepackage{epstopdf}

\definecolor{cream}{RGB}{222,217,201}

\usepackage[colorinlistoftodos]{todonotes}
\usepackage{color}

\usepackage{comment}

\begin{document}

\pagestyle{fancy}
\thispagestyle{plain}
\fancypagestyle{plain}{
\renewcommand{\headrulewidth}{0pt}
}

\makeFNbottom
\makeatletter
\renewcommand\LARGE{\@setfontsize\LARGE{15pt}{17}}
\renewcommand\Large{\@setfontsize\Large{12pt}{14}}
\renewcommand\large{\@setfontsize\large{10pt}{12}}
\renewcommand\footnotesize{\@setfontsize\footnotesize{7pt}{10}}
\makeatother

\renewcommand{\thefootnote}{\fnsymbol{footnote}}
\renewcommand\footnoterule{\vspace*{1pt}%
\color{cream}\hrule width 3.5in height 0.4pt \color{black}\vspace*{5pt}} 
\setcounter{secnumdepth}{5}

\makeatletter 
\renewcommand\@biblabel[1]{#1}            
\renewcommand\@makefntext[1]%
{\noindent\makebox[0pt][r]{\@thefnmark\,}#1}
\makeatother 
\renewcommand{\figurename}{\small{Fig.}~}
\sectionfont{\sffamily\Large}
\subsectionfont{\normalsize}
\subsubsectionfont{\bf}
\setstretch{1.125} 
\setlength{\skip\footins}{0.8cm}
\setlength{\footnotesep}{0.25cm}
\setlength{\jot}{10pt}
\titlespacing*{\section}{0pt}{4pt}{4pt}
\titlespacing*{\subsection}{0pt}{15pt}{1pt}

\fancyfoot{}
\fancyfoot[LO,RE]{\vspace{-7.1pt}\includegraphics[height=9pt]{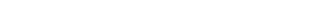}}
\fancyfoot[CO]{\vspace{-7.1pt}\hspace{13.2cm}\includegraphics{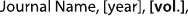}}
\fancyfoot[CE]{\vspace{-7.2pt}\hspace{-14.2cm}\includegraphics{head_foot/RF}}
\fancyfoot[RO]{\footnotesize{\sffamily{1--\pageref{LastPage} ~\textbar  \hspace{2pt}\thepage}}}
\fancyfoot[LE]{\footnotesize{\sffamily{\thepage~\textbar\hspace{3.45cm} 1--\pageref{LastPage}}}}
\fancyhead{}
\renewcommand{\headrulewidth}{0pt} 
\renewcommand{\footrulewidth}{0pt}
\setlength{\arrayrulewidth}{1pt}
\setlength{\columnsep}{6.5mm}
\setlength\bibsep{1pt}

\makeatletter 
\newlength{\figrulesep} 
\setlength{\figrulesep}{0.5\textfloatsep} 

\newcommand{\topfigrule}{\vspace*{-1pt}%
\noindent{\color{cream}\rule[-\figrulesep]{\columnwidth}{1.5pt}} }

\newcommand{\botfigrule}{\vspace*{-2pt}%
\noindent{\color{cream}\rule[\figrulesep]{\columnwidth}{1.5pt}} }

\newcommand{\dblfigrule}{\vspace*{-1pt}%
\noindent{\color{cream}\rule[-\figrulesep]{\textwidth}{1.5pt}} }

\makeatother

\twocolumn[
  \begin{@twocolumnfalse}
{\includegraphics[height=30pt]{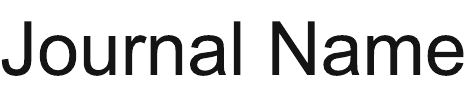}\hfill\raisebox{0pt}[0pt][0pt]{\includegraphics[height=55pt]{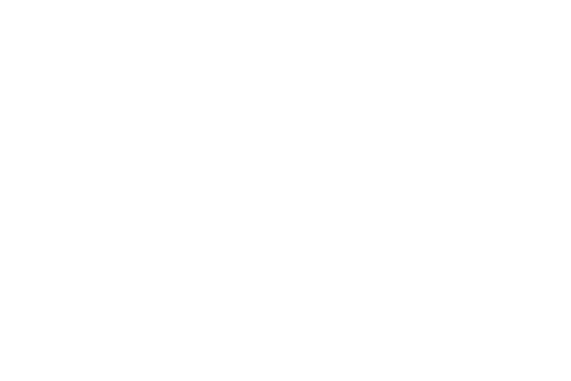}}\\[1ex]
\includegraphics[width=18.5cm]{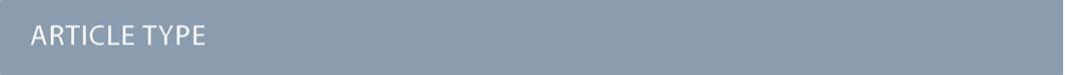}}\par
\vspace{1em}
\sffamily
\begin{tabular}{m{4.5cm} p{13.5cm} }

\includegraphics{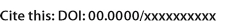} & \noindent\LARGE{\textbf{
Optimal Resource Utilization for Autonomous Laboratory Orchestrators
}} \\
\vspace{0.3cm} & \vspace{0.3cm} \\

& \noindent\large{ 
Austin McDannald$^{\ast}$\textit{$^{a}$}\textit{$^{b}$},
Julia Tisaranni\textit{$^{a}$\textit{$^{c}$}},
Howie Joress\textit{$^{a}$}\textit{$^{d}$}
}
 
 \\

\includegraphics{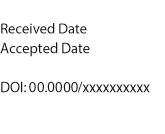} & \noindent\normalsize{
In autonomous laboratories, AI agents  suggest the next batch of experiments to do. 
However,  planning and executing those tasks  taking full advantage of the available resources is a completely different question. This can be challenging when dealing with real-world hardware constraints, especially so when there are multiple instruments with different capacities and throughputs. 
Here we demonstrate a 2-step method to address resource utilization for our autonomous platform for metal-organic framework synthesis. 
First, we use constraint programming to find optimal schedules. 
This finds schedules that minimizes the total time while still satisfying the limitations and capacities of the hardware. 
Secondly, we use a system of status dependencies for each task, which allows for the robust execution of the optimal schedules.} 

\end{tabular}

 \end{@twocolumnfalse} \vspace{0.6cm}

  ]

\renewcommand*\rmdefault{bch}\normalfont\upshape
\rmfamily
\section*{}
\vspace{-1cm}

\footnotetext{\textit{$^{a}$}Material Measurement Laboratory, National Institute of Standards and Technology, Gaithersburg, MD, USA}
\footnotetext{\textit{$^{b}$}ORCID: 0000-0002-3767-926X, austin.mcdannald@nist.gov}
\footnotetext{\textit{$^{c}$}ORCID: 0009-0008-0012-1714}
\footnotetext{\textit{$^{d}$}ORCID: 0000-0002-6552-2972}




\section{Introduction}

Autonomous systems are becoming ever more popular for experiments in materials and other physical sciences.\cite{Stach2021, Joress2024}
There is a recognized need to orchestrate the actions within an autonomous experimental platform.\cite{Joress2025} 
There have been a few developments of autonomous laboratory orchestration to date. 
These include ARES OS\cite{ARES_AM, ARES_OS}, HELAO-async\cite{HELAO-async}, NIMS-OS \cite{NIMS-OS}, ChemOS 2.0\cite{ChemOS, ChemOS_2_0}, EOS\cite{EOS}, Bluesky+ROS2\cite{Bluesky_ROS2}, MULTITASK\cite{Kusne2023}. 
Many of these efforts have been focused on facilitating the deployment of autonomous laboratory components: networking computational resources, instrument communication, implementing server-client models for the instrument AI actions, facilitating different architectures for how AI agents can interact between themselves and with the instruments. 
In this paper we address a slightly different problem, namely, given a set of requested experiments from the AI agents, how do we make optimal use of the available instrumental resources?

One of the issues with attempting to optimize the use and schedule of resources on an autonomous platform, is that we, as the operators and programmers, do not know \textit{a priori} what tasks will be asked of platform to execute.
In an autonomous system, there is an AI agent (or several \cite{Kusne2023}) that is attempting to answer some question at hand (e.g. find some optimal property, answer some scientific question). 
These AI agent(s), through their acquisition functions, are generating list of requested new data points. 
The orchestrator of the platform must make decisions about how those requests are executed: what actions are performed by what resource of the platform and when.
There are some cases when this orchestration might be trivial. 
For example, if there is only enough capacity to perform one job a time, then all the jobs must be executed serially. 
Also if there is enough capacity to perform all jobs in parallel, then the orchestration is also trivial. 
Lastly, if the platform can execute the jobs at the same speed that the AI sends requests, then there will only be 1 job in the list, which is again trivial. 
Complications can arise, however, when there is only capacity for some jobs in parallel - especially if those jobs can vary widely in execution time. 
Further complications can be when there are interrelated dependencies - e.g. parallel capacity, but only if certain conditions are met, or the execution of a task in one job depends on a task from another job.
In general simply finding optimal schedules for a set of jobs is NP-hard.\cite{Brailsford1999, Mnich2018}
Furthermore, orchestrating the execution of the schedules comes with its own challenges. 

In this work, we use the example of a robotic platform for the synthesis of metal-organic frameworks (MOFs). 
Some steps of the MOF synthesis require less than a few minutes (e.g. dispensing precursors), while other steps can take more than 12 hours (e.g. solvo-thermal reaction, drying).
Some of these steps have strong dependencies - the reaction cannot start before the precursor are added. 
Some of the resources also impose constraints - while the centrifuge can process several samples at a time, it is hard to add a sample to the centrifuge if it is already spinning. 

Finding optimal schedules within the constraints imposed by the physical components of the platform can be cast as a Job Shop Problem using a Constraint Satisfaction framework. 
In mathematics and computer science, Constraint Satisfaction Programming attempts to find the values of variables defined within a domain that satisfy all of the declared constraints. 
A Job Shop Problem is a common type of Constraint Satisfaction problem seen in operational research.\cite{Brailsford1999}
In Job Shop Problems there are a number of jobs that each need to be scheduled on specific machines, and the main variables are typically the start times, often with the goal of minimizing the total time to complete the entire set of jobs.
Job Shop Problems are often computationally complex,\cite{Mnich2018} however solvers have been developed that attempt practical solutions.
In this work, we use OR-Tools to find optimal schedules.\cite{OR-Tools}  
A discussion about the solver used in OR-Tools can be found here: \citealt{DominikKrupke2024}.

While finding optimal schedules is an important step in orchestrating the actions on the autonomous robotic platform, it is not sufficient for execution of the scheduled tasks.
In this paper we also introduce our system of status dependencies for the robust execution of tasks.
The rest of this paper is organized as follows: In Section 2 we introduce definitions that we use throughout this paper, in section 3 we provide details of our MOF synthesis platform, in Section 4 we discuss the framing and use of the scheduler, in Section 5 we discuss the system of status dependencies, and discuss the conclusions and open problems in the final section.

\section{Definitions}
For the sake of clarity, for the remainder of this article we will use the following definitions:

\begin{description}
    \item[] Campaign - A singular research effort containing multiple experiments all with a defined scientific goal and search space, e.g. discovering the synthesis landscape for a particular MOF. 
    \item[] Job - a job is an entire experiment, potentially consisting of many tasks and using several resources, e.g. complete synthesis of one sample.
    \item[] Task - some action within a job, typically only requiring the use of one resource. 
    \item[] UnitOP - short for unit operation, the function that executes a task.
    \item[] Resource - A component of the platform that gets used as part of the workflow.  Each resource has constraints related to its use, including the number of samples it can handle at one time.  
    \item[] Consumable - Items that a job uses that are single use (at least within the context of a campaign).  These include precursors (chemicals, reactants, and other types of feedstock), as well as other items that used in the reaction such as sample containers, pipette tips.

\end{description}

\section{Metal-Organic Framework Synthesis Platform}
The autonomous laboratory orchestration system described in this paper was developed out of a need to coordinate the actions of a robotic platform designed to synthesize MOFs via solvothermal synthesis.  One such example is the synthesis of Copper benzene-1,3,5-tricarboxylate (Cu-BTC MOF also known as HKUST-1, Cu\textsubscript{3}BTC\textsubscript{2}, Basolite C300, or MOF-99)\cite{wang2013metal}.  To contextualize the decisions and workflows used here we briefly describe the synthesis and how we carry it out.

Solvothermal synthesis has several steps.  
First the organic linkers (Trimesic acid for Cu-BTC) and metals salt (Copper nitrate) are dissolved into a mixture of solvents.  
In our case this premixing will be manual and the mixtures will serve as precursors for the platform, along with pure solvents for further dissolution.  The metal solution and linker solution are then mixed together along with additional solvents to reach their target concentration for each reaction.  The solution is then heated in a closed vessel for some amount of time, typically on the order of tens of hours.  
At this point MOF has been precipitated but must be "washed", in order to remove the remaining precursors and perform a solvent exchange, then dried.   
Finally the the MOF is dried and activated, typically by heating it, applying vacuum, or both.  

Our platform will be described elsewhere in detail but in short, the reactions and subsequent processing steps occur within a glass vial.
The platform consists of various stations with a robotic manipulator to move the vials between stations (along with some other tasks).
For convenience for tracking the status of both vials and resources, we assign each resource a 3-digit address.
The first digit is the type of resource: arm \& clamp, reactor, centrifuge, etc.
The next digit is which particular resource: reactor 3, or reactor 4, etc. 
And the last digit specifies the position in the resource.
We can track the location of the samples - or equivalently - the occupancy of each resource with these addresses. 
We can extend these addresses to also include the bank of syringe pumps, conveniently tracking which precursors are loaded into which syringe pumps. 
We further extend this address locations to account for new samples - a address with the first digit of 0 can refer to samples that have yet to be assigned to a vial. 

A synthesis job begins by selecting a vial from a rack of vials on the platform.  
The stock solutions are added to the vial by a bank of syringe pumps, each precursor being weighed to ensure accuracy of the addition.  
The vial is then closed and moved to one, of a number of, heater blocks, which has been preheated.  
The vial is then left in the heater block for the duration of its reaction.  Following the reaction the washing tasks begin.
The vial is moved to a centrifuge to compact the solid MOF material to the bottom of the tube.  
A needle is then used to draw off the excess solvent (i.e., the supernatant) and additional fresh solvent is added.  
The  vial is then placed  in the centrifuge to break up the MOF pellet and allow the solvent to fully wet the material.  
The vial is then left to rest in the vial rack at room temperature, letting it sit for several hours.  
This washing process is repeated several times (possibly with different solvents).  
Following the final wash step the drying process can begin.  
The drying process begins similarly to the washing process where the vial is centrifuged and as much solvent is removed by aspiration as possible.  
The vial is then placed in a heated reactor under vacuum and allowed to dry for several hours. 
Overall the complete synthesis for an individual sample can take on the order of a couple of days. 
However, many of the resources on the platform that are used with long dwell times are duplicated or capable of handling multiple samples: the vial rack has dozens of spots, there are 8 heating blocks, each with a capacity of 4 samples, the centrifuge holds up to 6 vials.  
This means the platform is capable of simultaneously capable of processing tens of samples in semi-parallel. 

In the case of our platform, most of the characterization (except for a balance that can provide MOF yield data and a camera that can provide MOF color) will be performed off of the platform on semi-automated tools that can handle batches of up to 16 samples.  
We expect, when working optimally, the turn around time for this feedback to be on the order of a day.

\section{The Scheduler}
Some schedulers, such as the Round-Robin scheduling algorithm commonly used for CPU's,\cite{Round_Robin, Scheduling-Book} can be tuned or configured to execute the next tasks when there is available capacity on the resources during the downtime of earlier tasks. 
While such a scheduler would certainty be advantageous over a naive sequential execution, it would not provide a method of expressing the constraints imposed by the hardware or the dependencies on tasks for other samples (e.g. reactions at the same temperature can be on the same reactor, but must end at the same time). 
Since we want to be able to receive updates to the list of experiments from the AI agents, we need a scheduler that is expressive enough to find optimal schedules that still satisfy all the relationships and dependencies between the tasks.

We cast the robot scheduling operation as a Job Shop Problem within the Constraint Satisfaction framework. 
For this, we break each requested synthesis job into several tasks. 
For each task, we prescribe what resources that task will use and its duration. 
For most tasks we can know \textit{a priori} which resources it will use: e.g. dispensing the precursors will occupy the syringe pumps. 
Likewise we know or can estimate the duration of most tasks with reasonable accuracy - dispensing the precursors will take roughly the same amount of time (within a few seconds) regardless of what composition is being dispensed. 
For some tasks, we may only know what \textit{type} of resource it will use - this is the case if there are several independent duplicates of the same resource. 
For example, our platform has several reactors, each with the capacity for 4 samples. 
Before scheduling we do not know which reactor will be used for the reactions of any sample, let alone what position will be used. 
The drying tasks also use the reactors, but the reactors can only perform the reaction task or the drying task at any given time, so there is a need to coordinate which tasks are performed by which reactors. 

In our example of autonomous MOF synthesis, the acquisition function provides a list of requested synthesis experiments - a list of jobs.
Each job can then be broken down into tasks, as shown in Table \ref{tab:tasks}.

\begin{table}[h]
\centering
\renewcommand{\arraystretch}{1.5}
\begin{tabular}{|l|l|l|l|}
\hline
\multicolumn{2}{|l|}{\textbf{Task}} & \textbf{Resource} & \textbf{Duration (min)} \\ \hline \hline
\multicolumn{2}{|l|}{dispense precursors} & arm \& clamp & 1 \\ \hline
\multicolumn{2}{|l|}{reaction} & reactor & $\in [30, 2880]$ \\ \hline

\multirow{5}{*}{\rotatebox{90}{wash cycle}} & centrifuge & centrifuge & 60 \\ \cline{2-4}
                                            & remove supernatant & arm \& clamp & 1 \\ \cline{2-4}
                                            & dispense solvent & arm \& clamp & 1 \\ \cline{2-4}
                                            & sonicate & sonicator & 20 \\ \cline{2-4}
                                            & hold & rack & 1440 \\ \hline

\multicolumn{2}{|l|}{dry} & reactor & 1440 \\ \hline
\end{tabular}
\caption{The set of tasks for each job (\textit{i.e.} the synthesis of one sample), the resources used for that task, and the duration. Note that the duration of the reaction task is one of the experiment variables, and therefor provided by the acquisition function of the autonomous agent in the science layer of the platform. Additionally, each synthesis might include several wash cycles perhaps with different solvents, compiled into a macro as indicated. The synthesis routine would escape the wash macro at the \texttt{remove supernatant} step after all the wash cycles have been completed.}
\label{tab:tasks}
\end{table}

Each of those tasks has a known (or estimated) duration, but variable starting times. 
We can now break down the list of jobs into a list of tasks. 
With this list of tasks we can impose constraints. 
For each sample, the tasks must be completed in order. 
The arm \& clamp has a capacity of 1. 
The sonicator has a capacity of 4. 
We can safely ignore the capacity of the rack - samples are created from the empty vials that are loaded onto the rack, so there will definitionally always be an open spot for any vial at hand. 
The centrifuge has a capacity of 6, but if any tasks overlap on the centrifuge then they must overlap entirely (the start times must be equal) - we cannot add a sample to the centrifuge if it is already spinning.  
We also want to avoid a scenario where the precursors are dispensed, but there is a long wait before the sample is moved to heater block to start the reaction. 
So we can additionally impose the constraint that the end of dispensing task must be within 4 minutes of the start of the reaction task for each sample. 
For the heater blocks the constraints are a bit more involved. 
There are several heater blocks, and we can use those for the reaction and drying tasks. 
Each heater block has a capacity of 4 - however the temperature of each slot on an individual block cannot be independently controlled. 
So if tasks overlap on the same heater block, then they must be at the same temperature.
For the solvo-thermal synthesis, each individual heater slot is designed with pressure seals, such that the solvent vapors are captured and pressure builds to the  vapor-pressure of the solvent mixture during the reaction. 
However, it is dangerous to release the pressure in a slot, so the entire block must be cooled to allow for recondensation of the solvent prior to retrieving a vial. 
Therefore, if the reaction tasks overlap on the same heater block, then they must \textit{end} at the same time. 
Our list of jobs can easily exceed the parallel capacity of the heater blocks. 
And while we can easily control heating rate of the blocks, they are not actively cooled.
So we can further impose the constraint that if two tasks use the same heater block at different times and temperatures, then the one with the lower temperature should start first. 
The drying tasks and reaction tasks use the reactors in different configurations.
We can impose that any drying tasks cannot overlap with any reaction tasks on the same reactor. 

To simplify the scheduling we can assign the reaction tasks to the reactors deterministically - as opposed to including those variables in the Job Shop problem. 
We can agglomerate reactions at the same temperature (up to the capacity of 4) to the same heater block - these reactions will be performed in parallel. 
We can further agglomerate the groups of tasks by appending the 2 shortest duration groups together, iteratively, until we are left with as many groups as reactors. 

Lastly, for the drying tasks, these can use any reactors and have no constraints (other than order of operations constraint already imposed) - so long as they do not overlap with a reaction task on the same reactor. 
To encode this within the constraint programming framework, we construct a boolean variables that effectively choose which reactor to use for drying, then impose the constraint that it does not overlap with the reaction tasks for that reactor. 

This is a decidedly sample-wise perspective breakdown of the jobs and tasks, which is convenient for keeping track of the data and metadata for the samples and ultimately the pertinent information for the AI agent attempting to answer some scientific or engineering question. 
However, this sample-wise treatment does have its limitations - e.g. for scheduling actions of resources that operate on multiple samples at once or on actions not tied to any one sample. 
Pre-heating the heater blocks is an example of this. 
We want the reactors to already be at the desired temperature when samples are loaded, and since there may be several hours between starting reactions of samples using the same reactor, it does not make sense to send new requests to heat the reactor for each sample as it is loaded. 
Instead, we interleave a preheating step that heats the whole reactor to the desired temperature. 
Then the react tasks simply load the samples into the preheated reactor, wait for the proper time, then unload the samples. 

Similarly, the centrifuge tasks do not fit well into this sample-wise breakdown of the tasks, since we do not want to send separate requests to start the centrifuge for each sample. 
This is handled by the function dependencies discussed in the next section, since this does not affect the scheduling. 

\begin{figure*}
    \centering
    \includegraphics[width=1.0\linewidth]{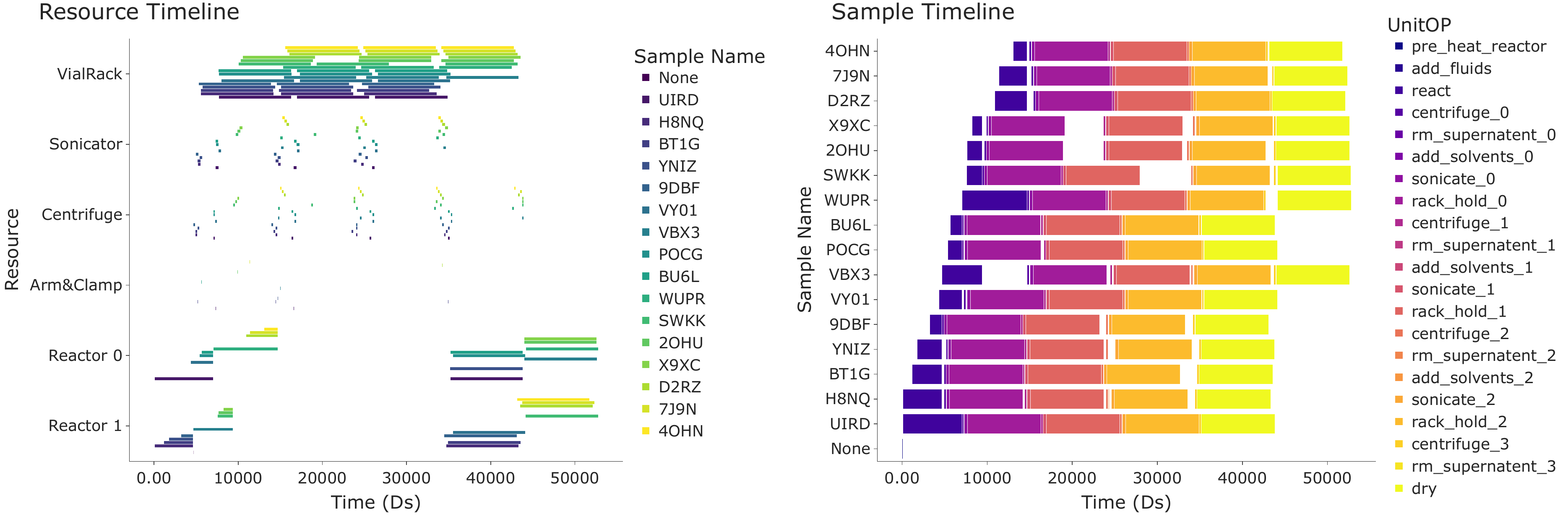}
    \caption{Gantt charts of the Tasks for the initial batch of Jobs in a Campaign. The left plot shows the tasks organized by which resource is used, while on the right the tasks are organized by sample. Note how the batches of reaction tasks on the same reactor and at the same temperature are constrained to end at the same time. At the plotted timescale the tasks involving the arm \& clamp are often too brief to see. As an aside, we chose decaseconds as the units for the variable times since the constraint satisfaction solvers operate on integers, minutes were too coarse resolution, and using seconds added unnecessary computational complexity.}
    \label{fig:gantt1}
\end{figure*}

The constraint satisfaction solver then attempts to discover schedules that minimize the total time to complete the list of jobs. 
One advantage to expressly encoding each of these constraints is the adaptability to new requests from the AI agents. 
The reaction and drying tasks will likely not conflict in a single batch of jobs - the react is at the beginning of each job, whereas the drying is at the end. 
Let's consider a concrete example.
In Fig. \ref{fig:gantt1} we have the first batch of Jobs in a Campaign.
This batch has 16 Jobs - with different reaction temperatures and times.
There are 4 sets of 4 samples, each set with its own reaction temperature, and each sample with its own reaction time. 
Each Job also has the remaining tasks from Table \ref{tab:tasks}, including 3 wash cycles. 
The scheduler finds optimal schedules that minimize the total time for the batch of Jobs, while still satisfying all the constraints.
Our scheduler, written using the Python package OR-Tools discovered the schedule shown in Fig. \ref{fig:gantt1} in about 28 s running on a 24 core CPU. 

\begin{figure*}
    \centering
    \includegraphics[width=1.0\linewidth]{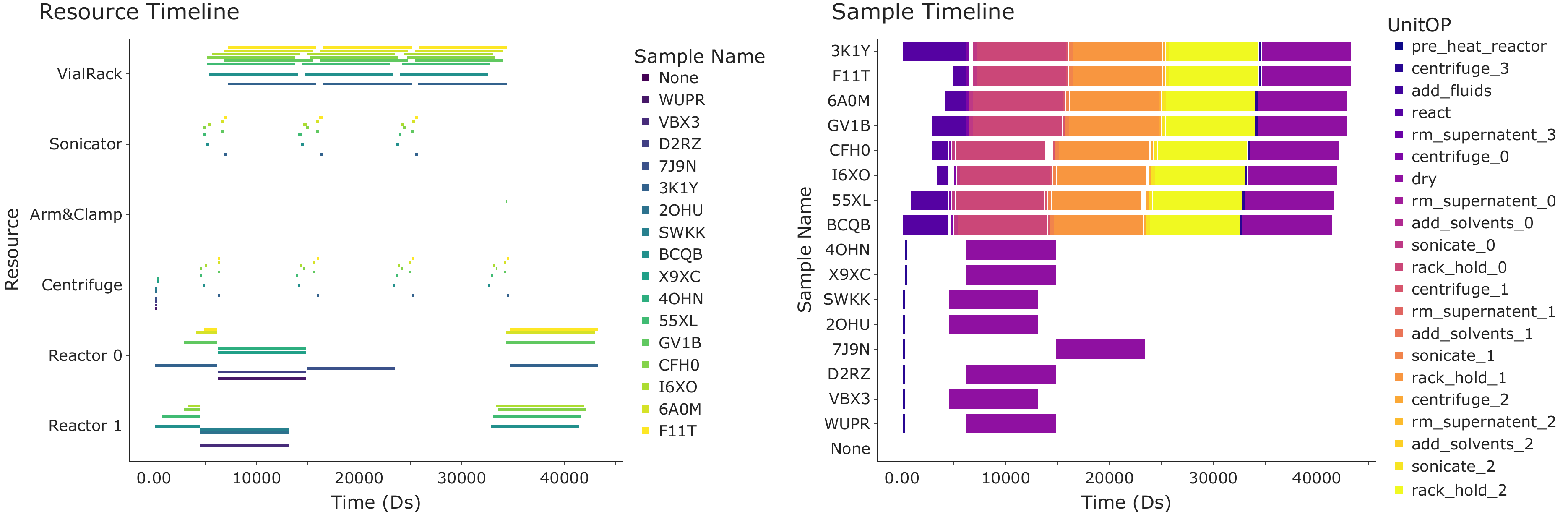}
    \caption{New Gantt charts of the Tasks after 8 new samples were introduced at 40000 Ds in the schedule from Fig. \ref{fig:gantt1}. The starting time for this set of tasks was reset to 0 Ds. The left plot shows the tasks organized by which resource is used, while on the right the tasks are organized by sample. Note how the drying tasks of the 8 samples from the previous batch can be optimally scheduled anytime there is open capacity between the end of the first reaction tasks and the start of the last drying tasks of the new 8 samples.}
    \label{fig:gantt2}
\end{figure*}

Now, let's suppose that the AI agent receives feedback on the experiments at 40000 Ds (aside: we used decaseconds so the integer variables have sufficient, but not superfluous fidelity), and uses that information to suggest the next batch of 8 samples. 
We want the scheduler to find optimal schedules for the remaining Tasks from the first batch of Jobs, as well as all the Tasks for new 8 Jobs.
This new schedule is shown in Fig. \ref{fig:gantt2}, which was discovered in about 1.4 s on the same 24 core CPU.
Note how none of the drying tasks overlap with the reaction tasks on the same reactor. 
Furthermore, since the reaction tasks are one of the early tasks in the sequence of tasks for each Job, scheduling any of the drying tasks at the beginning of experiment would delay the total completion time. 
From the perspective of the scheduler, which has the objective of minimizing the total completion time, the drying tasks from the previous 8 samples can be scheduled anytime there is open capacity on the reactors that doesn't interfere with the scheduling of the reaction or drying tasks from the new 8 samples.
That is, changing the start times any of the drying task from the previous 8 samples so that they fit in the window between about 4500 Ds and 34600 Ds has no impact on total reaction completion time, and is therefore considered equally optimal. 
Of course, in practice, we would like the previous 8 samples to be completed as soon as possible so that they could move on to characterization stages. 
Eventually this could be handled by increasing the scope of tasks modeled in the Job Shop scheduler.
For the time being, we handle this with the function dependencies (discussed in the next section) which executes the tasks as soon as status dependencies and component mutexes allow. 

\section{Function Dependencies}

The job shop scheduler is good for finding the order of tasks, and for estimating the start times and total duration, but it is not sufficient for the execution. 
The duration of the tasks in the scheduler are fixed and approximate - depending on these estimates could lead to unnecessary down time (if the estimate is too long) or conflicts (if the estimate is too short). 
Neither is the schedule detailed enough to use for execution. 
Moving the samples from one resource to another also requires the arm, but it would be too cumbersome to model that level of granularity in the job shop problem. 
This also means that the constraints imposed cannot be executed to the mathematical precision (reactions cannot end at the same time, since it takes non-zero time to, for example, move the sample from the reactor to the next resource). 

To robustly execute the list of tasks without strict reliance on the schedule and clocktime, we use a system of status dependencies and mutexes.
There is a mutex for every component of the system. 
If the component has capacity for more than one sample, there is an additional mutex for each position.
For example: there is a mutex for the arm \& clamp, but for a heater block with 4 sample positions there are 5 mutexes - 1 mutex for each position, and 1 mutex for heater block as a whole. 
Any function attempting to use any component must first checkout the mutex for that component - checking the status that the mutex is "Available" and changing that to "Occupied". 
For the reaction task as an example, the must checkout the mutex for an available position on the heater block, then checkout the mutex for the arm \& clamp. 
The arm can then move the sample to that position, and upon completion the function releases the mutex for the arm \& clamp but retains the mutex for the heater block position.
At the end of the reaction, the function must then checkout the mutex for the arm \& clamp once again to remove the sample from the heater block, and upon completion releases all mutexes. 
This allows for the arm to be free to perform other tasks while sample is reacting in the heater block. 

\begin{figure}
    \centering
    \includegraphics[width=0.5\linewidth]{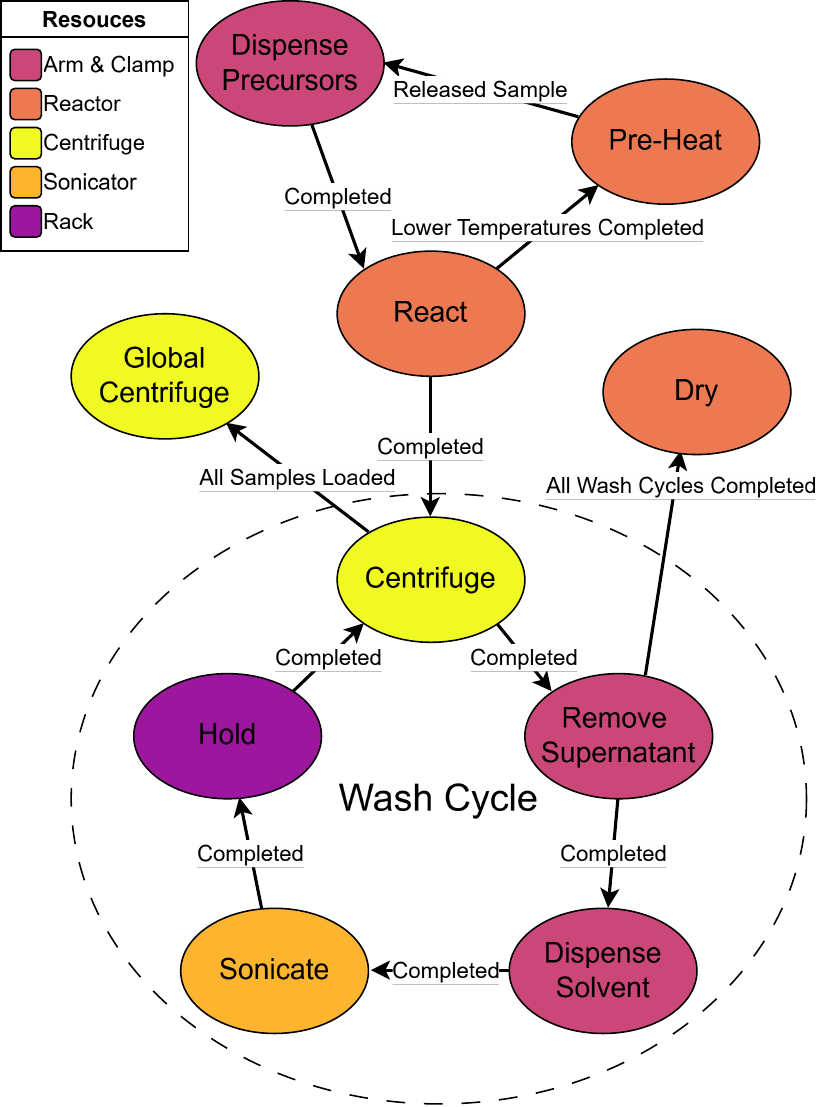}
    \caption{A simplified graph of the dependencies of the UnitOPs used for a MOF synthesis job. The color coding indicates the main resource each task. The arrows indicate what conditions should be met before proceeding to the next task.}
    \label{fig:dependency_graph}
\end{figure}

The UnitOP functions also have dependencies on the other tasks. 
A simplified diagram of the dependencies for the MOF synthesis job is shown in Fig. \ref{fig:dependency_graph}. 
We use a table as a centralized source detailing the status of each task for each sample.
For each MOF synthesis job, the first task is to dispense the precursors. 
This UnitOP depends on the \texttt{Pre-Heat} UnitOP for 2 reasons: first, the reactor that this sample will use might be processing an earlier set of samples, and second, as mentioned in the Job Shop Problem constraints, we do not want the precursors to be dispensed unless the reaction can be started within 4 minutes.
The \texttt{Pre-Heat} UnitOP depends only on any earlier reactions being completed.
So for the first set of samples scheduled, the \texttt{Pre-Heat} UnitOP starts warming the reactor. 
Once the reactor is at temperature, this UnitOP updates the status table to release the first sample to the \texttt{Dispense Precursors} UnitOP. 
Meanwhile the \texttt{Pre-Heat} UnitOP sleeps for the interval between the start times of the first and second samples, as determined by the scheduler, then releases the next sample, and so on until all samples have been released. 
The \texttt{Reaction} UnitOP execution depends on the precursor dispensing task being completed.
There is also a redundant check to ensure that the reactor is still at the desired temperature from the Pre-Heat task.
The sample-specific \texttt{Centrifuge} UnitOP only depends on the \texttt{Reaction} UnitOP - samples are ready to be centrifuged as soon as the reaction is complete. 
However, the centrifuge can hold several samples at a time. 
Similar to the \texttt{Pre-Heat} UnitOP, we use a separate UnitOP to send a single start command to the centrifuge, which we call the \texttt{Global Centrifuge} UnitOP. 
This \texttt{Global Centrifuge} UnitOP waits for each sample in the set to begin. 
The set of samples is provided by the scheduler. 
The \texttt{Centrifuge} UnitOP for each of those samples can update the status table to indicate that it is loaded into the centrifuge.
Once all the samples have been loaded, the \texttt{Global Centrifuge} UnitOP can check to make sure the centrifuge is balanced, then begin. 
Meanwhile each individual \texttt{Centrifuge} UnitOP must wait for the \texttt{Global Centrifuge} UnitOP to update the status, releasing the samples, an allowing them to be unloaded. 
The \texttt{Remove Supernatent}, \texttt{Dispense Solvent}, \texttt{Sonicate}, and \texttt{Hold} UnitOPs each simply depend on the previous UnitOP completing.
The \texttt{Dry} UnitOP depends on the last \texttt{Remove Supernatent} UnitOP for this sample. 

Combining the component mutexes and the dependencies allows for the robust control of the execution of all of the tasks from each of the jobs. 
A generic example of a UnitOP is shown in Alg. \ref{alg:GenericUnitOP}.
Our Python implementation of these algorithms make use of the AsyncIO package\cite{python_asyncio} for asynchronous execution of the tasks. 
In the UnitOPs, the order of the checks matter. 
The samples should be ready for the task before attempting to obtain the mutexes to the components for that task.
This means that the UnitOP should pass the dependency check before checking out the component mutexes. 
This order ensures that the UnitOP does hoard a resource before the sample is ready for that task, which might potentially create a situation where no UnitOPs are able to run and the entire autonomous campaign grinds to a halt.

\begin{table}
\begin{algorithm}[H]
    \caption{Generic Unit Operation}
    \begin{algorithmic}[1]
        \State \textbf{await} status\_dependencies\_check
        \State component\_mutex\_checkout:
            \State \hskip1.0em \textbf{await} component\_mutex == "Available"
            \State \hskip1.0em component\_mutex = "Occupied"
        \State perform task
        \State update task\_status
        \State release component\_mutex:
            \State \hskip1.0em component\_mutex = "Available"
    \end{algorithmic}
    \label{alg:GenericUnitOP}
\end{algorithm}
\caption*{The basic structure of a unit operation function. Note that the order of the checks matter. The function should pass all the status dependency checks before attempting to obtain the mutexes for components. This ensures that some task is not blocking the use of the components unnecessarily.}
\end{table}

The structure of a UnitOP for a task that uses a shared resource, as shown in Alg. \ref{alg:SharedResourceUnitOP} is even more nuanced. 
In addition to passing the dependency check before checking out the mutex needed component, these tasks should also check out the needed component before checking out the mutex for the arm \& clamp. 
This ensures that other UnitOPs can continue to execute while this UnitOP waits for the needed component to become available. 
As mentioned earlier, to avoid sending multiple commands for each sample-wise UnitOP that uses the shared resource, we use a separate UnitOP that directly controls this resource as a whole. 
There are 2 basic structures for these sample-wise UnitOPs: one for when the global component UnitOP is not blocking, and another for when it is blocking. 
For the \texttt{React} UnitOP, samples can be loaded and un-loaded from the reactors while other samples continue to use that reactor. 
Therefore the \texttt{Pre-Heat} UnitOP does not need block the execution of any of the \texttt{React} UnitOPs.
So the sample-wise \texttt{React} UnitOP, first moves the sample to the reactor, then simply sleeps for the duration of the reaction time, then moves the sample from the reactor. 
However, the \texttt{Global Centrifuge} UnitOP does need to be blocking, since samples can only be loaded or un-loaded when the centrifuge is stationary. 
Therefore the sample-wise \texttt{Centrifuge} UnitOP, moves the sample to the centrifuge, then waits for the \texttt{Global Centrifuge} UnitOP to update sample status to the appropriate condition before moving the sample from the centrifuge. 
The mutexes to the arm \& clamp are released after every move is completed to allow other UnitOPs to move samples and execute elsewhere on the platform. 

\begin{table}
\begin{algorithm}[H]
    \caption{Shared Resource Unit Operation}
    \begin{algorithmic}[1]
        \State \textbf{await} status\_dependencies\_check
        \State component\_mutex\_checkout:
            \State \hskip1.0em \textbf{await} component\_mutex == "Available"
            \State \hskip1.0em component\_mutex = "Occupied"
        \State arm\&clamp\_mutex\_checkout:
            \State \hskip1.0em \textbf{await} arm\&clamp\_mutex == "Available"
            \State \hskip1.0em arm\&clamp\_mutex = "Occupied"
        \State move\_to\_component
        \State release arm\&clamp\_mutex
        \State \textbf{IF} resource\_control is not blocking:\Comment{e.g. React UnitOP}
            \State \hskip1.0em \textbf{sleep} duration 
        \State \textbf{IF} resource\_control is blocking:\Comment{e.g. Centrifuge UnitOP}
            \State \hskip1.0em \textbf{await} status\_dependencies\_check
        \State \textbf{await} arm\&clamp\_mutex\_checkout
        \State move\_from\_component
        \State update task\_status
        \State release component\_mutex:
            \State \hskip1.0em component\_mutex = "Available"
        \State release arm\&clamp\_mutex:
            \State \hskip1.0em arm\&clamp\_mutex = "Available"
    \end{algorithmic}
    \label{alg:SharedResourceUnitOP}
\end{algorithm}
\caption*{The basic structure of a unit operation function that uses a shared resource. Here, again, the order of the checks matter. Firstly, the sample should be ready for this task. Then the mutex for the component of the shared resource should be obtained, and then the mutex for the arm\&clamp should be obtained. When the unit operation controlling the shared resource is not blocking, this sample-wise unit operation simply sleeps for the duration. When the unit operation controlling the shared resource is blocking, then this sample-wise unit operation waits for the appropriate status. The structure of this unit operation ensures that the execution of the function is appropriately blocking for the interplay of resources in the platform.}
\end{table}

The pertinent question now is: if we have this system of mutexes and dependencies do we still need the Job Shop scheduler? 
The answer is yes, but we don't rely on the start or stop times. 
From the scheduler, we use the order of samples on each reactor, and the time delay between the \texttt{Pre-Heat} steps and the \texttt{Dispense Precursors} steps so that each batch of samples on a reactor will finish their reactions as close to simultaneously as practically possible.
The rest of the UnitOPs would automatically execute an optimal schedule according to the constraints imposed by the mutex and dependencies. 
However, in the event of re-scheduling while some jobs are still incomplete - e.g. if new samples are requested by the AI agents - then the scheduler also ensures an order of operations that makes optimal use of resources within the constraints. 
For example the scheduler would ensure that no \texttt{React} UnitOPs from the new samples overlap with any of the \texttt{Dry} UnitOPs from earlier samples on the same reactor.

\section{Open Problems}

The Job Shop Scheduler finds schedules that minimize the total time. 
The system of status dependencies for the machines and the UnitOPs robustly executes those schedules - with the added benefit that jobs are completed as soon as possible (when there is capacity that does not impact the schedule). 
This would be an ideal solution to autonomous laboratory orchestration -- if all jobs were equally valuable. 

However, often the goal of autonomous laboratories is to accelerate the knowledge gain and there are a few open challenges that the framework presented here does not address: 

\textbf{1) Prioritization:} Often the acquisition functions of the AI agents generating the list new experiments to perform can additionally include some prioritization of those tasks - ranking or scoring the jobs based on predicted knowledge gain. 
The orchestrator presented here ignores that information; the scheduler takes advantage of efficiency gains due to batching, but only optimizes for the total completion time.

\textbf{2) Feedback:} Often the acquisition functions only consider the costs of each experiment independently. 
However, the time cost is not just the length of time required for each task for that sample, but is also affected by the other experiments that are part of that schedule. 
Furthermore, adding an experiment to the schedule will have knock-on effects to all subsequent experiments in that campaign.
Therefore, the true marginal time cost is only discernible after the entire campaign has been executed. 
Even limiting the scope of scheduling optimization to a single batch of jobs being scheduled, the scheduler is too computationally expensive to practically to feedback to the acquisition function. 

\textbf{3) Design of Experiments Decisions}: Because of points \textbf{1} and \textbf{2} above, there is limited ability for the scheduler to make higher-order design decisions about which experiments to perform.   
Given a list of jobs with forecasted expected knowledge gains for performing that experiment, the question for the scheduler is which experiments to run. 
The further into the forecasted list, the less accurate the forecasting will be, so those points may be less valuable.  
Should the platform start 8 experiments, 16 experiments, 24 experiments?  
The more that are started at one time, the faster data will be generated, but there is a higher likelihood the data will have less useful information. 
Conversely, starting fewer, but more informative, samples in a batch allows the next batch of experiments to start sooner, with those experiments having higher certainty of knowledge gain.

In the case of our platform and workflow, there is an even more difficult optimization problem that can be addressed.  
The ideal goal for an autonomous project is not to maximize the short term knowledge gain rate, but to maximize knowledge gain over a campaign (the length of which is typically bound by the researcher based on time, number of experiments, or amount of consumables). 
There is also no reason that experiments being queued need to be completed in order.  
In the case of our workflow, there is benefit of running, for instance, samples that are reacted at the same temperature together, because they can be run on a reactor together. 
While this type of clustering accelerates the rate of samples through the reaction stage (which is a large component of the synthesis time), completing the processing of these samples may slow the total flow of samples through the system.  
Again these samples may be less valuable and doing these syntheses may ultimately waste consumables.

Getting proper estimates on the value and costs (both in time and materials) of samples depends on what else is being run and therefore requires updating the forecast generated by the scientific agent as well as the scheduler.
We could, in principle, re-run the scheduler for each possible forecasted sample to accurately determine the overall costs, and repeat that as the acquisition function of the AI in the science layer adds each sample to populate the next batch of samples. 
However, this leads to a combinatorial explosion of an already computationally intensive task, which quickly becomes intractable. 

So, how many samples should be in the batch? 
And what samples should be in that batch - especially given efficiency gains from batching and speed of feedback from the results of the experiments?
Perhaps the answers to these questions changes as the research campaign progresses.
Solving this higher level orchestration problem is a quite important, but difficult, operations research problem that remains an open challenge.  



\section{Code Availability}

The code is available from \url{https://github.com/usnistgov/autoMOF}

\section*{Conflicts of Interest}
There are no conflicts to declare.


\section*{Disclaimer}
Any mention of commercial products in this report is for information only; it does not imply recommendation or endorsement by NIST.



\balance


\bibliography{rsc} 
\bibliographystyle{rsc} 

\end{document}